\documentclass[runningheads]{llncs}
\usepackage[T1]{fontenc}
\usepackage{graphicx}
\usepackage{amsmath,amsfonts}
\usepackage{threeparttable}
\usepackage{array}
\usepackage{tikz}
\usepackage{textcomp}
\usepackage{stfloats}
\usepackage{url}
\usepackage{verbatim}
\usepackage{graphicx}
\usepackage{cite}
\usepackage{caption}
\usepackage{subcaption}
\usepackage{amssymb}
\usepackage{xcolor}
\usepackage{booktabs}
\usepackage{makecell}
\usepackage[english]{babel}
\usepackage{enumitem}
\usepackage{braket}  
\usepackage{multirow}
\usepackage{booktabs}

\usepackage{bbding} 
\usepackage{algpseudocode, algorithm}
\usepackage{makecell}

\begin{document}
\title{Vehicle Routing Problems via Quantum Graph Attention Network Deep Reinforcement Learning}
\titlerunning{Quantum-GAT DRL Model for VRP}
\author{
Le Tung Giang\inst{1} \and 
Vu Hoang Viet\inst{2} \and 
Nguyen Xuan Tung\inst{3} \and 
\\ Trinh Van Chien\inst{2} \and 
Won-Joo Hwang \inst{1,} \textsuperscript{\Envelope} 
} 
\authorrunning{T.G. Le  et al.}
%
\institute{School of Computer Science and Engineering, Pusan National University, \\ Busan, 46241, Republic of Korea \\
\email{\{giang.lt2399144, wjhwang\}@pusan.ac.kr} \\
\and School of Information and Communications Technology, \\Hanoi University of Science and Technology, Hanoi, Vietnam\\
\email{viethoangvu0808@gmail.com, chientv@soict.hust.edu.vn}
\and Faculty of Interdisciplinary Digital Technology, PHENIKAA University, \\Yen Nghia, Ha Dong, Hanoi, 12116, Viet Nam\\ \email{tung.nguyenxuan@phenikaa-uni.edu.vn}
}
\maketitle              
\begin{abstract}
The vehicle routing problem (VRP) is a fundamental NP-hard task in intelligent transportation systems with broad applications in logistics and distribution. Deep reinforcement learning (DRL) with Graph Neural Networks (GNNs) has shown promise, yet classical models rely on large multi-layer perceptrons (MLPs) that are parameter-heavy and memory-bound. We propose a Quantum Graph Attention Network (Q-GAT) within a DRL framework, where parameterized quantum circuits (PQCs) replace conventional MLPs at critical readout stages. The hybrid model maintains the expressive capacity of graph attention encoders while reducing trainable parameters by more than $50\%$. Using proximal policy optimization (PPO) with greedy and stochastic decoding, experiments on VRP benchmarks show that Q-GAT achieves faster convergence and reduces routing cost by about $5\%$ compared with classical GAT baselines. These results demonstrate the potential of PQC-enhanced GNNs as compact and effective solvers for large-scale routing and logistics optimization.

\end{abstract}

\vspace{-2mm}
\section{Introduction}
\vspace{-2mm}

Intelligent Transportation Systems (ITS) are a key component of smart cities \cite{2019_TITS_Survey}, yet e-commerce growth has made delivery costs a major burden. The vehicle routing problem (VRP) is a well-known NP-hard problem in combinatorial optimization and has been studied extensively for decades, having wide applications such as express delivery \cite{PERBOLI201919}, and production planning \cite{2019_TITS_Logistic}, highlighting their practical importance.
Varius heuristic and meta-heuristic solvers yield approximations but rely on good initializations and suffer due to the complexity as instance size grows \cite{1989_tabu} .

To circumvent the real-time inference challenge, deep reinforcement learning (DRL) has been applied to NP-hard VRP \cite{20}. DRL can learn patterns beyond handcrafted heuristics and provides fast inference for large-scale, time-sensitive VRPs \cite{2019_TITS_DRL}. However, classical neural models often ignore the graph topology, limiting flexibility and generalization. Graph Neural Networks (GNNs) address this remedy by exploiting topology \cite{TUNG_survey} and excel in combinatorial tasks \cite{2024_HGNN, 2025_TVT_Tung}. However, their common multi-layer perceptron (MLP) readouts are often parameter-heavy and memory-bound.

With the exploration of quantum computing, Quantum Machine Learning (QML) has emerged as a promising approach. QML addresses this via parameterized quantum circuits (PQCs) that capture high-order correlations with far fewer parameters \cite{2024_Mark}. Prior studies show that QML reduces model complexity and accelerates convergence \cite{2025_TVT_Hieu, 2025_Giang_ICAIIC}. Motivated by this, we propose a DRL model for graph-structured VRP instances, introducing a Quantum Graph Attention Network (Q-GAT) encoder that enhances feature learning, reduces model complexity, and accelerates convergence. The main contributions are as follows:

\begin{itemize}
    \item  We develop a quantum enhanced DRL framework with graph attention network (GAT) encoder, where parameterized quantum circuits (PQCs) replace MLPs in the meta-layer to enhance message-passing capacity. 
    \item We train the Q-GAT model with proximal policy optimization (PPO), an actor–critic algorithm supporting both greedy and stochastic decoding.  
    \item We validate Q-GAT on various VRP benchmarks, showing its ability to capture complex combinatorial dependencies beyond classical GNN baselines.  
    \item Experiments confirm that Q-GAT reduces parameters by over $50\%$, converges faster, and improves routing performance by about $5\%$, marking one of the first applications of PQC-enhanced GNNs to logistics and supply chain optimization.  
\end{itemize}

\vspace{-2mm}
\section{Vehicle Routing Problem}
\vspace{-2mm}

\begin{figure}[t]
	\centering
	\includegraphics [width=0.65\linewidth]{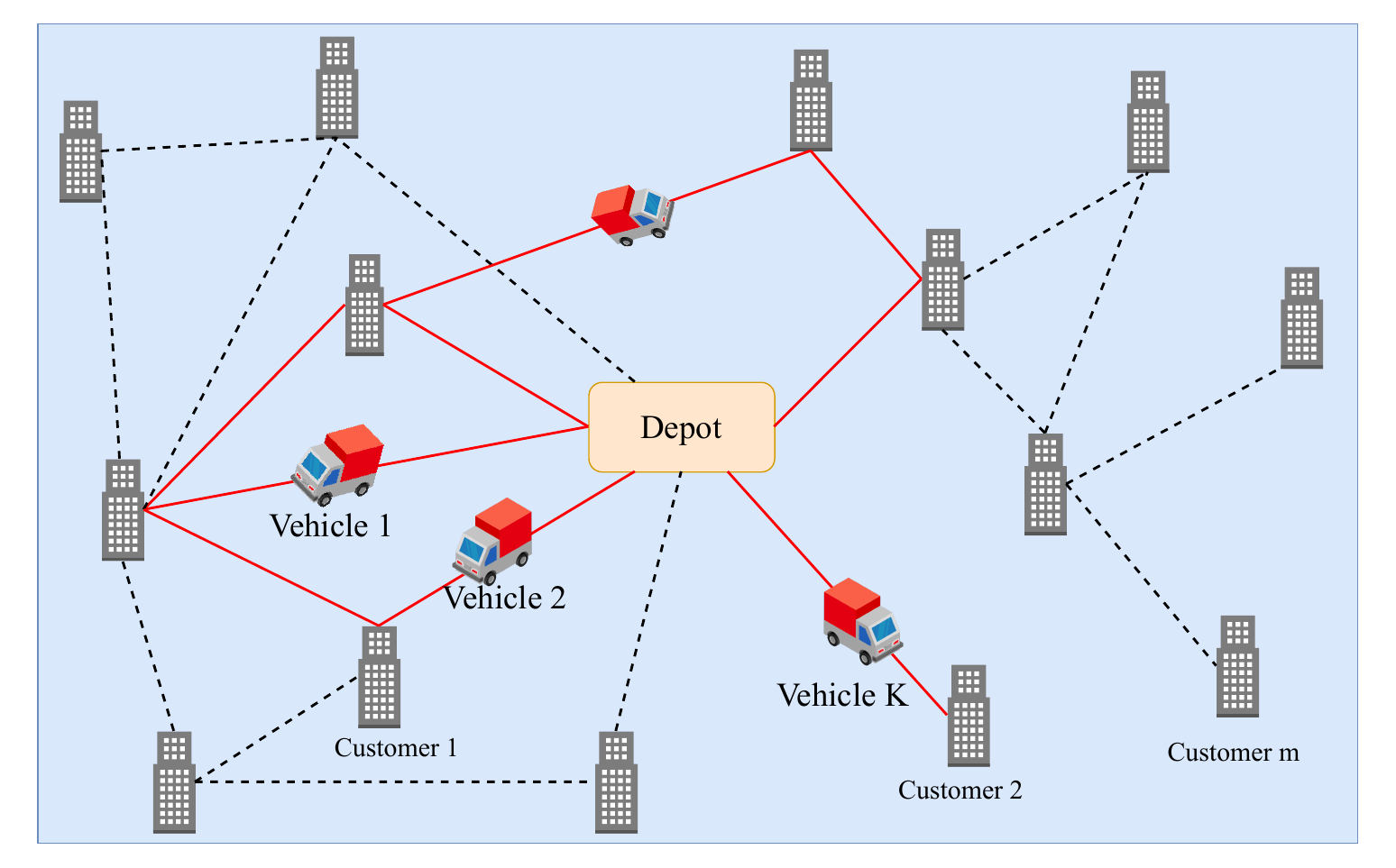}
	\caption{The architecture of the model}
	\label{fig:vrp}
    \vspace{-0.75cm}
\end{figure}
A capacitated VRP instance specifies a depot, a set of customers with demands, and a fleet of identical vehicles of limited capacity.
We consider the VRP on an undirected graph $G=(\mathcal{V},\mathcal{E})$  consisting of a set of nodes $\mathcal{V}=\{0,1,\ldots,m\}$, and a set of edges $\mathcal{E} = \{(i, j) \mid i < j, \; i, j \in \mathcal{V}\}$.
We denote node $0$ as the depot, and the customer set $\mathcal{V}_{c}=\mathcal{V} \setminus \{0\}$. Each customer node holds the feature including its coordinate $\mathbf{n}_{i}$, and its demand $d_{i}$ with $0<d_{i}$.
Assume that $K$ identical vehicles have capacity $C$, each starts from and ends at the given depot. The decision variable $x_{ij}$ equals $1$ if arc $(i,j)$ belongs to the optimal route, and $0$ otherwise, it is not included in the optimal route. 
Given $\pi$ as the solution route for a problem of graph $G$, we interested in the total routing length:
\begin{align}\label{eq:length}
    L(\pi \vert G) = \sum_{i=1}^{\vert\pi\vert-1} \vert \vert \mathbf{n}_{\pi(i)} - \mathbf{n}_{\pi(i+1)}  \vert \vert_{2},
\end{align}
where $\vert\vert \cdot\vert\vert_{2}$ is the $l_{2}$ norm and $\vert\pi\vert$ is the length of the route.

The objective is to design a set of minimum length routes such that each customer is served exactly once, all routes start and end at the depot, and the total demand on each route does not exceed the vehicle capacity.  As depicted in Fig.~\ref{fig:vrp}, several vehicles are needed to
meet the customer demand according to the following requirements:
\begin{subequations}\label{eq:vrp}
\begin{align}
\underset{\pi}{\textrm{minimize}} \quad& L(\pi \vert G) \label{eq:vrp-obj} \\ 
\textrm{s.t.} \qquad
        & x_{ij}\in\{0,1\},  \forall (i,j)\in \mathcal{E}. \label{eq:vrp-binary}\\
        &\sum_{i\in V} x_{ij} = 1, \quad  \forall j \in \mathcal{V}_{c} \label{eq:vrp-in}\\
        & \sum_{j\in V} x_{ij} = 1, \quad  \forall i \in \mathcal{V}_{c} \label{eq:vrp-out}\\
        & \sum_{j\in V} x_{0j} = K,\quad \sum_{i\in V} x_{i0} = K \label{eq:vrp-depot}\\
        & u_i \ge d_i, \quad \forall i \in \mathcal{V}_{c} \label{eq:vrp-load}\\
        & u_i - u_j + C\,x_{ij} \le C - d_j, \quad \forall i\neq j,\ i,j\in \mathcal{V}_{c} \label{eq:vrp-capacity}
\end{align}
\end{subequations}
Here, objective \eqref{eq:vrp-obj} minimizes the total length of the optimal route;
constraint \eqref{eq:vrp-binary} defines the binary routing decision variable;
constraints \eqref{eq:vrp-in} and \eqref{eq:vrp-out} ensure that each customer is visited exactly once;
constraint \eqref{eq:vrp-depot} enforces that each vehicle departs from and returns to the depot;
constraint \eqref{eq:vrp-load} ensures the cumulative load on the route passing the $i$-th customer; and constraint \eqref{eq:vrp-capacity} guarantee that the cumulative demand on any route does not exceed the vehicle capacity $C$ and eliminate subtours disconnected from the depot.  
We further define the stochastic policy based on the chain rule of probability:
\begin{align} \label{eq:pro}
    p_{\theta}(\pi \vert G) = \prod_{t=1}^{m}p_{\theta}(\pi_{t} \vert G, \pi_{<t}),
\end{align}
where $\theta$ are the policy parameters and $\pi_{<t}$ denotes the partial sequence up to step $t-1$.
To solve the problem \eqref{eq:vrp}, we use a DRL model to compute every component in the right-hand side of \eqref{eq:pro} and the basic elements of the DRL model are drfined as follows

\begin{itemize}
    \item \textbf{Agent:} Each vehicle is an agent. At time step $t$, it observes the environment state and selects an action $\pi_{t}$ according to its policy network. After executing the action, it receives a scalar reward and updates parameters to improve routing decision.
    \item \textbf{State:} The state comprises a static part and a dynamic part.
    The static state contains coordinates for VRP, which does not change during decoding. The dynamic state captures: remaining customer demands $d_i^{(t)}$, remaining vehicle load $C^{(t)}$, and position of the operating vehicle at current time $t$.
    \item \textbf{Action:}
The action \(\pi_t\) specifies the next destination at step \(t\). A complete joint action sequence \(\pi\) defines the route set for the VRP. The episode terminates when all demands are satisfied, yielding a feasible VRP solution.
    \item \textbf{Reward:}
We optimize routing cost by rewarding the agent with the negative objective in \eqref{eq:length}. 
\end{itemize}

\vspace{-2mm}
\section{Deep Reinforcement Learning Model}
\vspace{-2mm}

\begin{figure}[t]
	\centering
	\includegraphics[width=0.9\columnwidth]{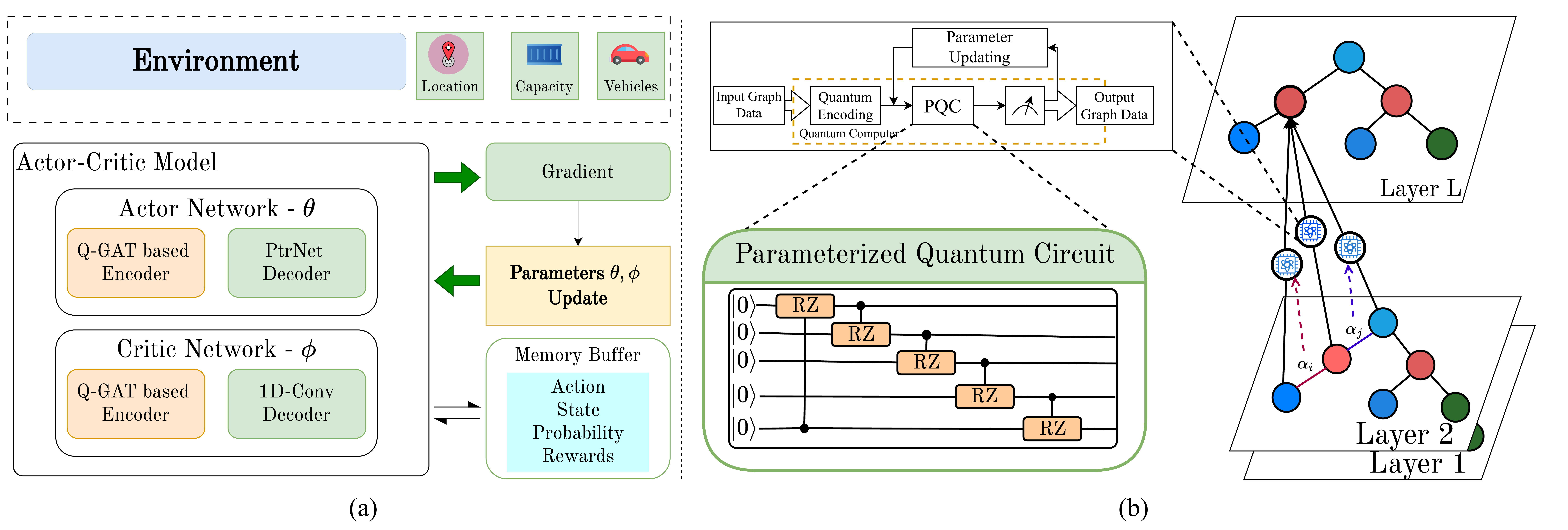}
	\caption{(a) The architecture of the DRL model. (b) The Quantum Graph Attention Network framework.}
	\label{fig:model}
    \vspace{-0.5cm}
\end{figure}

In this section, we proposed a DRL model to solve the aforementioned VRP instance, as illustrated in Fig.~\ref{fig:model}(a). In particular, the model is based on the actor-critic algorithm, where each network is consists of the sharing Q-GAT-based encoder, and a specific decoder. The actor network employ a pointer network that sequentially select the next node, while the critic network employ a 1D-convolutional readout network for expected reward. 

\subsection{Quantum Graph Attention Network-based Encoder}

We design a quantum-enhance graph attention network (GAT) for the encoder, where each MLP is replaced with QNN model. The detailed framework of Q-GAT model is depicted in Fig.~\ref{fig:model}(b). This replacement may improve the feature learning with high-dimension Hilbert in quantum space, and reduce the model complexity, leading to a speed up in training convergence.

\subsubsection{Message Passing Layer with Edge Attention}

Let $n_i$ be the raw feature of node $i$ and $e_{ij}$ the feature of edge $(i,j)$. 
We first compute initial node and edge embeddings as:
\begin{align}
x_i^{(0)} &= BN(\boldsymbol{A}_0 n_i + b_0), \quad \forall i\in V, \label{eq:node-embed}\\
\hat{e}_{ij} &= BN(\boldsymbol{A}_1 e_{ij} + b_1), \quad \forall (i,j)\in A, \label{eq:edge-embed}
\end{align}
where $BN(\cdot)$ is batch normalization, and $A_0,A_1$ are learnable input processing parameters.
At each layer $\ell$, the attention coefficient between node $i$ and $j$ is:
\begin{equation}
a_{ij}^{(\ell)} = \frac{\exp\!\left(\sigma\!\left(g^{(\ell)T} W^{(\ell)} [x_i^{(\ell-1)} \,\|\, x_j^{(\ell-1)} \,\|\, \hat{e}_{ij}]\right)\right)}
{\sum_{z=1}^m \exp\!\left(\sigma\!\left(g^{(\ell)T} W^{(\ell)} [x_i^{(\ell-1)} \,\|\, x_z^{(\ell-1)} \,\|\, \hat{e}_{iz}]\right)\right)},
\label{eq:attention}
\end{equation}
where $\sigma(\cdot)$ is a LeakyReLU to facilitate nonlinearity and $\|$ denotes concatenation.  
In conventional GAT, the inner transformation $W^{(\ell)}(\cdot)$ is implemented via a learnable layers. The node embeddings are then updated using a residual connection:
\begin{equation}
x_i^{(\ell)} = \sum_{j=1}^m a_{ij}^{(\ell)} W_1^{(\ell)} x_j^{(\ell-1)} + x_i^{(\ell-1)}.
\label{eq:residual-update}
\end{equation}
After $L$ layers of attention, we obtain the final node representations $x_i^{(L)}$. A global graph representation is derived by average pooling:
\begin{equation}
\bar{x}_j = \frac{1}{m} \sum_{i=1}^m (x_i^{(L)})_j, \quad j=1,\ldots,d_x,
\label{eq:graph-embed}
\end{equation}
where $d_x$ is the embedding dimension.

\subsubsection{Quantum Neural Network}
To enhance the message passing learning, we replace the classical MLPs in \eqref{eq:attention}-\eqref{eq:residual-update} by parameterized quantum neural networks (QNNs).
Each QNN performs three operations:

\noindent$\bullet$ Quantum Embedding:
Classical features $z \in \mathbb{R}^d$ are mapped into a quantum state by an encoding unitary $U_\text{enc}(z)$:
\begin{equation}
    | \psi(z) \rangle = U_\text{enc}(z) |0\rangle^{\otimes n}.
\end{equation}

\noindent$\bullet$ Parameterized Quantum Circuit (PQC)
The quantum state is processed by a trainable PQC consisting of rotation and entangling gates, parameterized by $\theta$:
\begin{align}
    | \psi_\theta(z) \rangle = U_\text{pqc}(\theta) \, | \psi(z) \rangle.
\end{align}

\noindent$\bullet$ Measurement
The output embedding is obtained by measuring expectation values of observables $O$:
\begin{align}
    h(z) = \langle \psi_\theta(z) | O | \psi_\theta(z) \rangle.
\end{align}
This measurement vector $h(z)$ serves as the updated node/edge feature and is passed to the next message-passing step. The replacement of QNN allows better feature learning and increases the convergence speed. 

\subsection{The Decoder}

\subsubsection{Pointer Network}
The decoder in actor network follow a pointer network (PtrNet) structure \cite{PtrNet} with Transformer-style multi-head attention to sequentially construct a tour. At the first decoder layer, we initialize the attention coefficient for each node as below:
\begin{align}
u^{(1)}_{i,t} =
\begin{cases}
\displaystyle \frac{q^\top k_i}{\sqrt{d_v}}, & i \notin \{\hat{\pi}_{t'}\}_{t'<t},\\[2mm]
-\infty, & \text{otherwise},
\end{cases}
\end{align}
In this step, we mask out every selected node by setting the coefficient to $-\infty$, then we normalize all coefficients via a softmax activation function
\begin{align}
    \hat{u}^{(1)}_{i,t} = \mathrm{softmax}_i\!\left(u^{(1)}_{i,t}\right).
\end{align}
Then, the context vector is calculated through a fully connected layer:
\begin{align}
c^{(1)}_t = W_f \left( \big\|_{h=1}^H \sum_{i=1}^m \hat{u}^{(1,h)}_{i,t} v^{(h)}_i \right),
\end{align}
where $W_f$ is the learnable weight matrix. The multi-head attention mechanism enhances the stability of attention learning. This context vector is passed through the second decoder layer to obtain the attention coefficient as follows
\begin{align}
u^{(2)}_{i,t} &=
\begin{cases}
\displaystyle C\,\tanh\!\left(\frac{(c^{(1)}_t)^\top k_i}{\sqrt{d_v}}\right), & i \notin \{\hat{\pi}_{t'}\}_{t'<t},\\[2mm]
-\infty, & \text{otherwise},
\end{cases}
\end{align}
Then, the probability for each customer node is obtained through a final softmax activation function:
\begin{align}
p_{i,t} = \mathrm{softmax}_i\!\left(u^{(2)}_{i,t}\right).
\end{align}
This probability distribution is then process via decoding strategy to predict the next node to visit.
In this work, we use the following decoding strategies:

\noindent$\bullet$ \textbf{Greedy Decoding:} 
A greedy algorithm selects locally optimal choices at each step, offering a fast approximation to the global optimum. In each decoding step, the node with the highest probability is chosen, and all visited nodes are masked out. For the VRP, termination occurs when the demands of all nodes are satisfied, thereby constructing a feasible solution.

\noindent$\bullet$ \textbf{Stochastic Sampling:} 
At each decoding step $t \in \{1, \dots, m\}$, the random policy 
$p_\theta(\hat{\pi}|s, \hat{\pi}_{t'} \, \forall t' < t)$ samples the next node according to the learned probability distribution. To ensure diversity, in \cite{2021_KunLei}, authors used a temperature hyperparameter $k \in \mathbb{R}$ to modify the probability distribution as
\begin{equation}
    p_{i,t} = p_\theta(\hat{\pi}|s, \hat{\pi}_{t'} \, \forall t' < t) = \text{softmax}\!\left(\frac{u_{i,t}}{k}\right).
\end{equation}
In \cite{2021_KunLei}, authors note that $k = 2.5,\, 1.8,\, 1.2$ are most effective for VRP instances of the same sizes. 


During the training phase, stochastic sampling is typically employed to sufficiently explore the solution space and improve the generalization capability of the model. In contrast, during the testing phase, we adopt greedy decoding to generate deterministic solutions with higher stability and reproducibility.

\begin{algorithm}[t]
\footnotesize
\caption{PPO for Combinatorial Optimization}
\label{al:PPO_Al}
\begin{algorithmic}[1]
\Require epochs $E$; data-collection steps per epoch $T_c$; PPO update epochs $K$; mini\-batch size $B$; actor $\pi_\theta$; critic $v_\phi$; clip $\varepsilon$; loss weights $c_v, c_e$
\State \textbf{Init} actor/critic parameters $(\theta,\phi)$; empty memory buffer $\mathcal{M}$
\For{$e = 1,\dots,E$}
  \State $\mathcal{M} \leftarrow \varnothing$
  \For{$t = 1,\dots,T_c$}
    \State Sample a batch of problem instances $\mathcal{S}$
    \State Roll out $\pi_\theta$ on $\mathcal{S}$ to obtain trajectories $\tau$ (states, actions, log-probs)
    \State Compute per-instance rewards and store tuples in $\mathcal{M}$
  \EndFor
  \State Normalize rewards in $\mathcal{M}$
  \State Compute advantages $\hat{A}$ using critic $v_\phi$ 
  \For{$k = 1,\dots,K$}
    \For{each mini-batch $\mathcal{B}\subset\mathcal{M}$ of size $B$}
      \State Compute probability ratios $r(\theta)$ 
      \State Actor loss: clipped surrogate $\mathcal{L}_{\text{CLIP}}(\theta)$
      \State Value loss: $\mathcal{L}_v(\phi)$ on returns vs.\ $v_\phi(s)$
      \State Entropy bonus: $\mathcal{L}_e(\theta)$
      \State Total loss: $\mathcal{L} = -\mathcal{L}_{\text{CLIP}}(\theta) + c_v\,\mathcal{L}_v(\phi) - c_e\,\mathcal{L}_e(\theta)$
      \State Update $\theta,\phi$  to minimize $\mathcal{L}$
    \EndFor
  \EndFor
\EndFor
\State \textbf{Output:} trained actor parameters $\theta$
\end{algorithmic}
\end{algorithm}

\subsection{Proximal Policy Optimization Training Algorithm}
In this section, we present a proximal policy optimization (PPO) \cite{PPO} to train our model.
The PPO framework following the actor and critic networks \cite{2021_KunLei} for routing problem is illustrated in Fig.~\ref{fig:model}. In particular, the Q-GAT model serves as the actor network, while the critic network is composed of multiple QNNs layer and a one-dimensional convolutional layer to output a scalar value $v_{\phi}(s)$. This value is used to estimate the cumulative rewards, which is defined as the total travel length in \eqref{eq:length}. In the PPO,  parameter updates are performed after an entire episode has been collected. Given the critic value, we define the advantage estimate as:
\begin{align}
    \hat{A} = L(\pi,s) - v_{\phi}(s),
\end{align}
To train the actor network, we use the clipped surrogate objective loss function $\mathcal{L}_{\text{CLIP}}(\theta)$ and the policy entropy loss function $\mathcal{L}_{\text{E}}(\theta)$ \cite{PPO} defined as follows:
\begin{align}
    \mathcal{L}_{\text{CLIP}}(\theta) &= \mathbb{E}\left[ \min\left( r(\theta)\hat{A}, \; \text{clip}(r(\theta), 1-\epsilon, 1+\epsilon)\hat{A} \right)\right],\label{eq:ppo-clip}\\
    \mathcal{L}_{\text{E}}(\theta) &= \text{Entropy}(p_\theta(\hat{\pi}|s)),\label{eq:entropy}
\end{align}
where $\epsilon$ is the clipping coefficient and $r(\theta)$ is the probability ratio between the updated and the previous policy:
\begin{align}
    r(\theta) = \frac{p_{\theta}(\pi\vert s)}{p_{\theta'}(\pi\vert s)}.
\end{align}
The clipped surrogate objective $\mathcal{L}_{\text{CLIP}}(\theta)$ \eqref{eq:ppo-clip} constrains each policy update by bounding the probability ratio between new and old policies \cite{PPO}. This prevents excessively large updates under stochastic returns and thereby stabilizes training of the actor \cite{PPO}. The policy entropy $ \mathcal{L}_{\text{E}}(\theta)$\eqref{eq:entropy} regularizes the objective by encouraging action diversity, improving exploration and reducing premature convergence to suboptimal deterministic behaviors. Jointly, these components balance stability and exploration, enhancing the actor’s sample efficiency and robustness on combinatorial routing tasks.
The critic network is trained by the Mean Squared Error (MSE) loss:
\begin{align}\label{eq:mse}
    \mathcal{L}_{\text{MSE}}(\phi) = \text{MSE}(L(\hat{\pi}|s), \hat{v}_\phi(s)).
\end{align}
Then the total loss function of the actor-critic model is expressed as:
\begin{align}
    \mathcal{L}(\theta, \phi) = \lambda_p \mathcal{L}_{\text{CLIP}}(\theta) + \lambda_v \mathcal{L}_{\text{MSE}}(\phi) - \lambda_e \mathcal{L}_E(\theta),
\label{eq:ppo-loss}
\end{align}
where $\lambda_{p}, \lambda_{v}, \lambda_{e}$ are the loss function weights balancing policy, value, and entropy terms. The detail of PPO algorithm is formulated in Algorithm~\ref{al:PPO_Al}.

        

\vspace{-2mm}

\section{Numerical Results}
\vspace{-2mm}

In this section, we conduct extensive simulation to evaluate the performance of the Q-GAT DRL model in solving the VRP task.

\subsubsection{Simulation setup.}
\textcolor{black}{
We trained the proposed Q-GAT DRL model on VRP instances with $m\in\{20,50,100\}$ customers. All training, validation, and random test sets were generated i.i.d. in the unit square $[0,1]\times[0,1]$ under the same distribution across splits. The demand of a customer point is a discrete number $d_i \in \{1, \ldots, 9\}$ chosen uniformly random. We set the maximize load as $(m, C) \in \{(20, 30), (50, 40), (100, 50)\}$. 
All models were trained for $100$ epochs  with a batch size of $256$ using the Adam optimizer and a learning rate of  $10^{-4}$. Each validation and random test set contained $10,000$ instances drawn from the same distribution.  For performance comparison, we consider both model-based and learning-based benchmarks such as sequence-to-sequence PtrNet~\cite{2018_RL}, solver LKH3~\cite{LKH3}, classical GAT~\cite{2021_KunLei} and Google’s OR-Tools.
}
\subsubsection{Training Evaluation.}
\begin{figure}[t]
	\centering
	\includegraphics[width=0.7\linewidth]{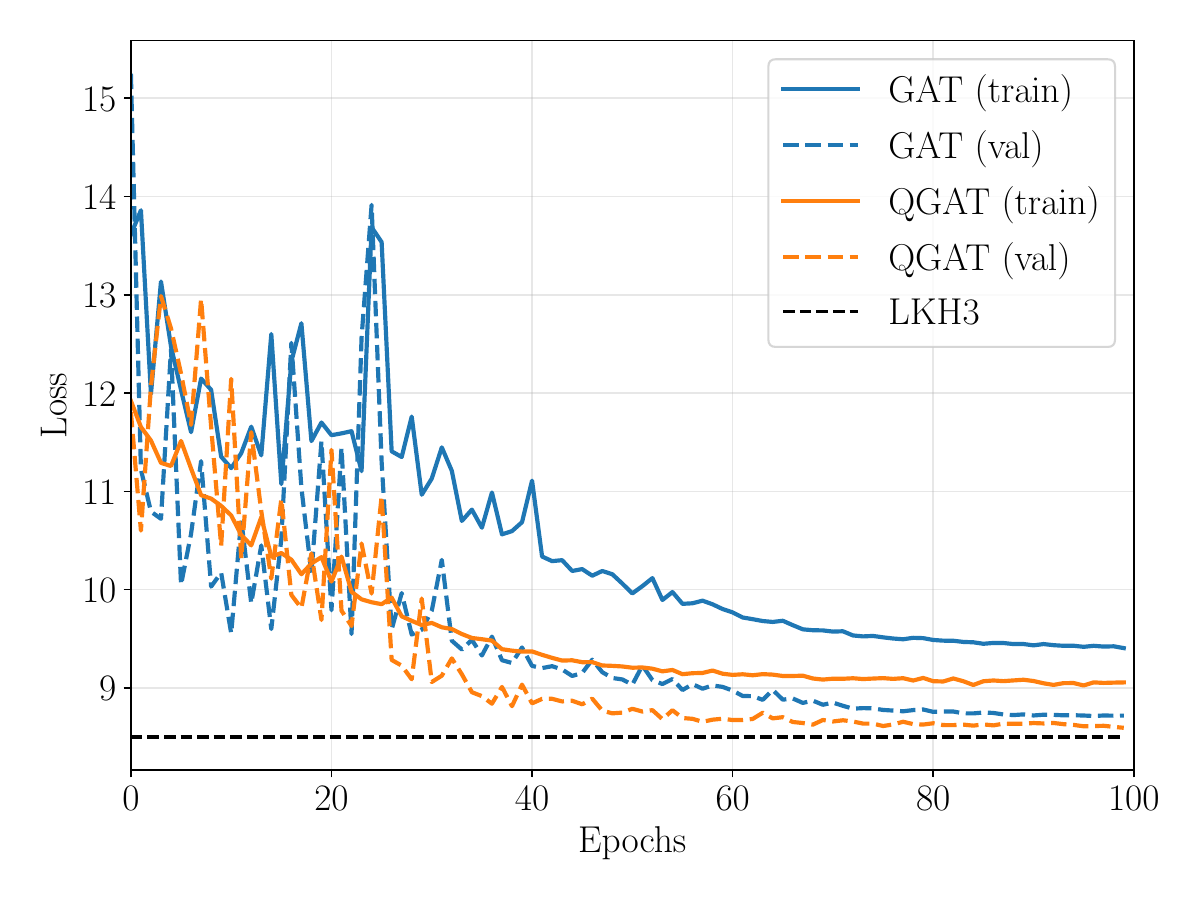}
	\caption{Training and testing loss versus the number of epochs with $20$ customers.}
	\label{fig:vrp20_loss}
    \vspace{-0.5cm}
\end{figure}

Figure~\ref{fig:vrp20_loss} shows the training and validation loss trajectories for both learning‑based models. Loss decreases rapidly in both cases, indicating stable optimization; however, Q‑GAT descends more steeply and levels off at a lower plateau than the classical GAT. For Q‑GAT, the train–validation gap becomes small after roughly $20$ epochs, whereas GAT exhibits larger oscillations and requires about $40$ epochs to stabilize. At convergence, Q‑GAT lies closer to the LKH3 benchmark. Overall, in this setting, Q‑GAT converges faster and attains higher final solution quality than the classical GAT.

\subsubsection{Model Complexity.}

\begin{table}[t]
  \centering
  \renewcommand{\arraystretch}{1}
  \setlength{\tabcolsep}{1em} 
  \caption{Model complexity comparison of Q-GAT and classical GAT.}
  \label{tab:param_counts}
  \begin{tabular}{lccc}
    \toprule
    \midrule
    \textbf{Model} & \makecell{\textbf{Classic}\\\textbf{parameters}} & \makecell{\textbf{Quantum}\\\textbf{parameters}} & \makecell{\textbf{Total Trainable}\\\textbf{parameters}} \\
    \midrule
    Classical GAT        & 324,493 & 0        & 324,493 \\
    Q-GAT       & 153,137 &  1,350   & 154,487 \\
    \midrule
    \bottomrule
  \end{tabular}
  \vspace{-0.75cm}
\end{table}
Table~\ref{tab:param_counts} shows that, while classical GAT utilizes $324,493$ trainable parameters, Q-GAT uses only $154,487$ trainable parameters, reducing $52.4\%$ model complexity. This is due to the light-weight PQC replacing the MLPs in classical model. The PQC message passing block requires only $1,350$ rotation angle parameters, while achieving the better performance.

\subsubsection{Scalability Evaluation.}
\begin{table}[t]
  \centering
  \renewcommand{\arraystretch}{1}
  \begin{threeparttable}
    \caption{Performance comparison on CVRP instances.}
    \label{tab:performance}
    \setlength{\tabcolsep}{4pt}
      \begin{tabular}{ll|cc|cc|cc}
        \toprule
        \multicolumn{2}{c|}{} &
          \multicolumn{2}{c|}{VRP20} &
          \multicolumn{2}{c|}{VRP50} &
          \multicolumn{2}{c}{VRP100} \\
        Method & Type &
          Length & Gap &
          Length & Gap &
          Length & Gap \\
        \midrule
        LKH3 & Solver & 8.54 & 0.00\% & 11.54 & 0.00\% & 20.82 & 0.00\% \\
        \midrule
        PtrNet   & RL, G & 9.22 & 8.02\% & 12.67 & 9.80\% & 25.13 & 10.12\% \\
        \textbf{Q-GAT (Greedy)}   & \textbf{RL, G} &
          \textbf{8.81} & \textbf{3.21\%} &
          \textbf{11.97} & \textbf{3.73\%} &
          \textbf{22.20} & \textbf{6.63\%} \\
        GAT (Greedy)             & RL, G & 8.98 & 5.13\% & 12.01 & 4.07\% & 22.21 & 6.67\% \\
        OR-Tools                  & H, S  & 8.95 & 4.81\% & 12.58 & 9.01\% & 22.83 & 9.65\% \\
        PtrNet    & SL, S & 8.91 & 4.33\% & 12.40 & 7.45\% & 22.57 & 8.40\% \\
        \textbf{Q-GAT (Sampling)} & RL, S  &
          \textbf{8.72} & \textbf{2.08\%} &
          \textbf{11.82} & \textbf{2.42\%} &
          \textbf{21.49} & \textbf{3.22\%} \\
        GAT (Sampling)           & RL, S  & 8.87 & 3.85\% & 11.92 & 3.29\% & 21.50 & 3.26\% \\
        \bottomrule
      \end{tabular}%

    \begin{tablenotes}[flushleft]
      \footnotesize
      \item \textbf{Note:} 
       In the Type column, the abbreviations are defined as follows: H denotes heuristic methods, SL refers to supervised learning, S indicates sample search, G represents greedy search.
    \end{tablenotes}
  \end{threeparttable}
  \vspace{-0.75cm}
\end{table}

Table~\ref{tab:performance} demonstrate the performance of both learning and model-based methods, in compared with the baseline of LKH3 solver.
Across all VRP instance sizes, Q-GAT achieves the smallest gap among learning-based methods under both greedy and stochastic decoding. With greedy decoding, Q-GAT consistently outperforms PtrNet and classical GAT. On average, it narrows the gap to LKH3 by about $3.5\%$ compared with PtrNet and by $0.77\%$ compared with classical GAT, highlighting the efficiency of the quantum-enhanced model even with a simple decoding strategy. With stochastic sampling, for $m=50$, it achieves a tour length of $11.82$, corresponding to a $2.42\%$ gap, which improves over GAT by $0.87\%$ and outperforms PtrNet in both RL and SL settings. Notably, for $m=100$, Q-GAT reduces the gap to $3.22\%$, compared with $3.26\%$ for GAT and $8.40\%$ for supervised PtrNet. These results demonstrate that the Q-GAT provides consistent improvements in both decoding modes, yielding higher solution quality than purely classical GAT DRL architectures.
\vspace{-2mm}
\section{Conclusion}
\vspace{-2mm}
In this paper, we proposed a quantum–classical DRL framework for solving the VRP. By introducing a Q-GAT encoder, where PQCs replace conventional MLP readouts, we achieved a more compact and expressive model. The Q-GAT encoder preserves the advantages of graph attention while reducing the total number of trainable parameters by $50\%$. Extensive experiments on VRP benchmarks demonstrated that the proposed approach converges faster and improves solution quality by about $5\%$ compared with classical GAT baselines. These results confirm the effectiveness of integrating quantum computation into graph learning for combinatorial optimization. Beyond VRP, the hybrid design is general and can be extended to other routing and scheduling tasks in intelligent transportation and logistics. This work provides one of the first practical demonstrations of PQC-enhanced GNNs for optimization, opening promising directions for future research on scalable hybrid solvers.

\vspace{-0.5cm}
{\color{black}\subsubsection{Acknowledgements} 
This work was supported by Quantum Computing based on Quantum Advantage challenge research(RS-2024-00408613) 
through the National Research Foundation of Korea (NRF) funded by the Korean government (Ministry of Science and ICT (MSIT))
This research was supported by Creation of the quantum information science R\&D ecosystem(based on human resources) (Agreement Number) through NRF funded by the Korean government ((MIST)) (RS-2023-00256050).
This work was supported by the NRF grant funded by the Korea government (MSIT) (RS-2024-00336962)
Following are results of a study on the “Busan Regional Innovation System \& Education (RISE)” Project, supported by the Ministry of Education and Busan Metropolitan City
}
\vspace{-0.5cm}

\bibliographystyle{splncs04}
\bibliography{reference}
%
\end{document}